\pdfoutput=1

\documentclass[11pt]{article}

\usepackage[final]{acl}

\usepackage{times}
\usepackage{latexsym}

\usepackage[T1]{fontenc}

\usepackage[utf8]{inputenc}

\usepackage{microtype}

\usepackage{inconsolata}

\usepackage{graphicx}

\usepackage{booktabs}
\usepackage{float}

\title{Code-Driven Inductive Synthesis: Enhancing Reasoning Abilities of Large Language Models with Sequences}

\author{
 \textbf{Kedi Chen\textsuperscript{1}\thanks{This work was done during the internship at Shanghai AI Laboratory.}},
 \textbf{Zhikai Lei\textsuperscript{1}},
 \textbf{Fan Zhang\textsuperscript{3}},
 \textbf{Yinqi Zhang\textsuperscript{1}},
 \\
 \textbf{Qin Chen\textsuperscript{1}},
 \textbf{Jie Zhou\textsuperscript{1}},
 \textbf{Liang He\textsuperscript{1}},
 \textbf{Qipeng Guo\textsuperscript{2}},
 \textbf{Kai Chen\textsuperscript{2}},
 \textbf{Wei Zhang\textsuperscript{1}\thanks{Corresponding author.}}
\\
 \textsuperscript{1}East China Normal University,
 \textsuperscript{2}Shanghai AI Laboratory,
 \textsuperscript{3}Georgia Institute of Technology
 \\
 \{kdchen,kausal\}@stu.ecnu.edu.cn \{qchen, jzhou, lhe\}@cs.ecnu.edu.cn, \\
 \{zhang.inch,zhangwei.thu2011\}@gmail.com, \{guogipeng,chenkai\}@pjlab.org.cn, fanzhang@gatech.edu
}

\begin{document}
\maketitle
\begin{abstract}
Large language models make remarkable progress in reasoning capabilities.
Existing works focus mainly on deductive reasoning tasks (e.g., code and math), while another type of reasoning mode that better aligns with human learning, inductive reasoning, is not well studied. 
We attribute the reason to the fact that obtaining high-quality process supervision data is challenging for inductive reasoning.
Towards this end, we novelly employ number sequences as the source of inductive reasoning data.
We package sequences into algorithmic problems to find the general term of each sequence through a code solution.
In this way, we can verify whether the code solution holds for any term in the current sequence, and inject case-based supervision signals by using code unit tests. 
We build a sequence synthetic data pipeline and form a training dataset CodeSeq.
Experimental results show that the models tuned with CodeSeq improve on both code and comprehensive reasoning benchmarks.
\end{abstract}

\section{Introduction}
Recent advances in AI, including openai-o1 \citep{DBLP:journals/corr/abs-2409-18486} and deepseek-r1 \citep{deepseekai2025deepseekr1incentivizingreasoningcapability} make remarkable progress in reasoning capabilities of large language models (LLMs) \citep{DBLP:journals/corr/abs-2309-07864, DBLP:journals/fcsc/XuCPZXZWZWC24,DBLP:journals/tkde/JinLHJJH24,DBLP:journals/corr/abs-2304-00008}, such as mathematical reasoning \citep{DBLP:conf/eacl/AhnVLLZY24,DBLP:journals/corr/abs-2403-00896} and code reasoning \citep{DBLP:conf/nips/LiuXW023,DBLP:journals/tosem/JiangDWFSLJJ24}.


Existing works focus mainly on deductive reasoning tasks (e.g., code and math) \citep{DBLP:conf/iclr/WangRZLLSZSZ024,DBLP:journals/corr/abs-2410-08196}, utilizing general principles and axioms to logically achieve specific conclusions.
In contrast, another mode of reasoning, inductive reasoning \citep{DBLP:journals/cogsr/HanRPK24}, involves drawing general conclusions from specific patterns. 
This paradigm is key to knowledge generalization and better aligns with human learning.
However, limited research are conducted in this area.

We attribute the reason to the fact that obtaining high-quality process supervision data \citep{DBLP:journals/corr/abs-2403-04642} is quite challenging. 
In math-type problems, each step of the derivation process can be annotated and verified \citep{yang2024qwen25mathtechnicalreportmathematical}. 
However, the intermediate steps in inductive reasoning are relatively open, making it difficult to determine correctness. 
This leads to challenges in data construction and, consequently, hardness in model learning. 

\begin{figure}[t]
  \includegraphics[width=\columnwidth]{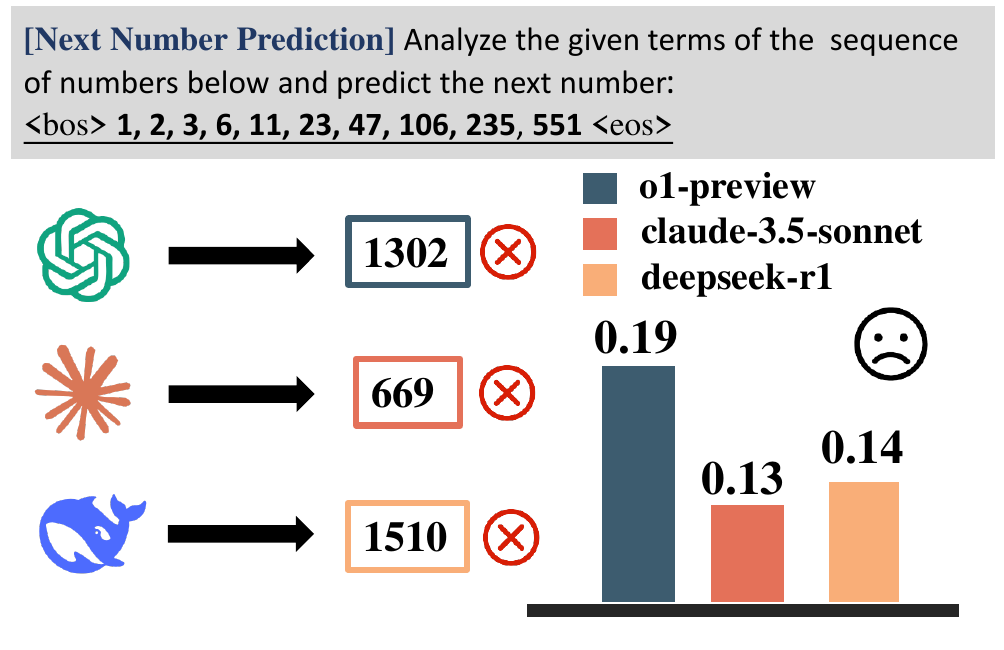}
  \caption{We select 200 sequences and prompt three powerful models for next number prediction (more details in Appendix~\ref{app:next number}). The results demonstrate that existing LLMs perform poorly in inductive reasoning, indicating significant research potential in this area.}
  \label{fig:sequence}
\end{figure}

In this paper, we novelly employ number sequences as the source of inductive reasoning data.
Sequence problems require generalizing from previous observations to predict future elements, which can reflect the inductive ability (see Figure~\ref{fig:sequence}).
We package sequences into algorithmic problems to find the general term of each sequence through a code solution.
In this way, we can verify whether the code solution holds for any term in the current sequence, and inject case-based supervision signals via code unit tests \citep{hui2024qwen25codertechnicalreport}.
Sepcifically, we build a sequence synthetic data \citep{DBLP:journals/corr/abs-2401-02524} pipeline guided by code unit tests, then forming a training dataset \textbf{CodeSeq}.

The pipeline consists of three steps.
\textbf{(1) Data filtering}. We scrape many sequences and their related information from websites. 
We use manually written rules and a language model working agent to filtrate the sequences that have enough information to be packaged into algorithmic problems.
\textbf{(2) Problem generation}. We leverage the working agent to generate an algorithmic problem about the general term for each selected sequence, along with two example cases. 
Another guiding agent directly generates the output based on the problem description and the input of example cases to verify whether the algorithmic problem itself is correct.
\textbf{(3) Supervision injection}. The working agent generates code solutions for the correct problems.
We verify whether the code solution holds for any term in the sequence through code unit tests.
The guiding agent provides modification suggestions and asks the working agent to regenerate the answers for the failed solutions.
Through this pipeline, we inject case-based supervision signals while searching for general term code solutions for sequences, forming the complete synthetic dataset CodeSeq.

To verify the effectiveness, we apply it to perform supervised fine-tuning (SFT) on two LLMs. 
Experimental results show that the models tuned with CodeSeq improve on two code benchmarks and three comprehensive reasoning benchmarks.

Our contributions can be listed as follows:
\begin{itemize}
\item{To our knowledge, we are the first to utilize sequences as the inductive reasoning data and study their impact on LLMs.}
\item{We package the sequences into algorithmic problems, which can be injected with case-based supervision signals to improve data quality for the inductive reasoning task.}
\item{Our synthetic data CodeSeq is proven effective for various reasoning tasks, demonstrating the potential of inductive reasoning.}
\end{itemize}

\begin{figure*}[t]
  \includegraphics[width=\linewidth]{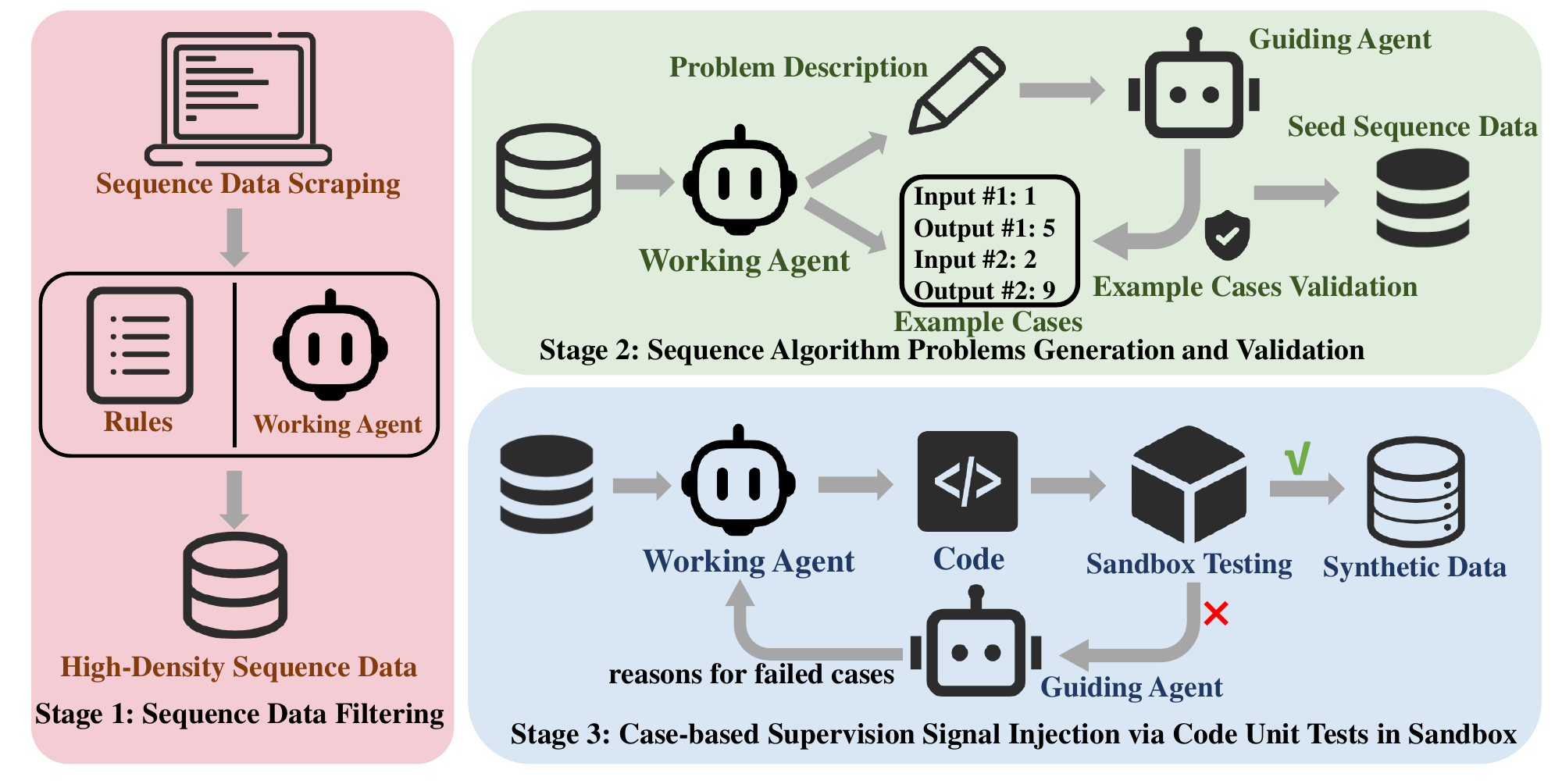}
  \caption{The sequence synthetic data pipeline consists of three steps, and then forming our CodeSeq.}
  \label{fig:pipline}
\end{figure*}

\section{Sequence Synthetic Data Pipeline}
In this paper, we employ sequences as the source of inductive reasoning data.
We package sequences into algorithmic problems to find the general term of each sequence through a code solution.
In this way, we can verify whether the code solution holds for any term in the current sequence, and inject case-based supervision signals by using code unit tests. 
The whole pipeline consists of three steps in Figure~\ref{fig:pipline}.
More details can refer to Appendix~\ref{app: pipeline}.

\subsection{Sequence Data Filtering}
We scrape a large number of sequences and their related information from websites\footnote{https://oeis.org/}. 
Each page on the website corresponds to a sequence and all its information, including the source, formula, general term description, and so on.

We will package the sequences into algorithmic problems by a powerful language model working agent. 
To ensure the accuracy of this process, we need to filter the information for each candidate sequence.
We first manually wrote rules to filter out sequences with insufficient information, such as those with too few terms, or those that evolve from other sequences (requiring additional webpage links for reference).
Then we prompt the working agent to self-planning \citep{jiang2024selfplanningcodegenerationlarge} the steps for generating an algorithmic problem and self-reflecting \citep{DBLP:conf/acl/WangZ0MZ024} on whether each step contains enough information.
The above operations result in a batch of sequences with high information density.

\subsection{Sequence Algorithmic Problem Generation and Validation}
We next have the working agent generate an algorithmic problem about the general terms for each sequence, along with two example cases. 
Example cases provide the standard input and output cases for this algorithmic problem to help the problem solvers understand it better.

To further verify the correctness of the algorithmic problems, we utilize another powerful LLM as a guiding agent. 
We input the problem description and two example cases' inputs into it and let it directly output the results. 
By comparing these results with the ground truth outputs generated by the working agent, we can determine whether the current problem is correct.
Seed sequence data is gained via this example case validation. 

\subsection{Case-based Supervision Signal Injection}
After obtaining the seed data, we let the working agent directly generate the code solution for the algorithmic problem. 
Since the problem description involves the general term of a sequence, the code solution represents the computational process for the general term of the sequence.
Unlike the example cases, we also set 5 to 7 test cases for each sequence to ensure the correctness of the code solution.

Imitating previous unit tests \citep{hui2024qwen25codertechnicalreport}, we use test cases to test the correctness of each code solution in an isolated sandbox environment.
If a code solution fails on a test case, we ask the guiding agent to provide the reason for the failure. 
We then give that reason along with the test case back to the working agent to correct the code solution.
Ultimately, through continuous self-correcting \citep{DBLP:conf/iclr/0009CMZYSZ24}, we achieve a code solution that passes all the test cases.

\subsection{Synthetic Data Statistics}
Based on the above process, we record the code of the current version each time a modification is made and generate synthetic data for each sequence, then forming a training dataset CodeSeq.
The data organization details of CodeSeq are provided in the Appendix~\ref{app:statistical}.

To ensure the diversity of the training data, we perform resampling \citep{DBLP:conf/emnlp/HirotaAZPMNX24} on the problem descriptions and the initially generated code solutions.
This operation is equivalent to resetting the starting point of the reasoning data, thereby obtaining a richer training corpus.
We use LLaMA3-8B model as the tokenizer and the final data statistics of CodeSeq can be found in Table~\ref{tab: statistical}.
From the table, we can see that our CodeSeq has a rich set of tokens available for training with an average of about 3 correction rounds. 
This proves that we effectively incorporate supervision signals into the sequence inductive reasoning data.

\begin{table}[ht]
\centering
\fontsize{9pt}{9pt}\selectfont
\renewcommand{\arraystretch}{1.5} 
\begin{tabular}{cc}
\toprule
Sample Form                     & SFT form                 \\
Sample Numbers                  & 9242                     \\
All Tokens                      &  15.3M                   \\
Output Tokens                   &  9M                     \\
Output Max Tokens               &  4273                    \\
First Hit Rate                  &  0.52                    \\
Avg Correction Rounds           &  2.93                    \\
Max Correction Rounds           &  5                       \\
\bottomrule
\end{tabular}
\caption{The data statistical information of CodeSeq. `First Hit Rate' indicates the probability that the first-generated code can pass all test cases.}
\label{tab: statistical}
\end{table}

\section{Experiments}
To prove the effectiveness of our sequence inductive reasoning synthetic data CodeSeq, we employ it to perform SFT on existing LLMs. 
We test its performance on code and other comprehensive reasoning benchmarks. We also explore whether CodeSeq could enhance the models' inductive reasoning capabilities.

\subsection{Training, Benchmarks, and Evaluation}
We conduct SFT on two widely used LLMs: LLaMA3-8B \citep{grattafiori2024llama3herdmodels} and Qwen2.5-7B \citep{qwen2025qwen25technicalreport}.
To maintain the models' instruction-following ability \citep{DBLP:conf/acl/0001ZZCXLWD24}, we mix CodeSeq with the latest post-training \citep{DBLP:conf/acl/WilliamsA24} 
corpus Tulu3 \citep{lambert2025tulu3pushingfrontiers} for SFT.
We then test the tuned models on two code benchmarks: Humaneval \citep{chen2021codex} and MBPP \citep{austin2021programsynthesislargelanguage}, along with three comprehensive reasoning benchmarks: MMLU \citep{hendrycks2021measuringmassivemultitasklanguage}, BBH \citep{suzgun2022challengingbigbenchtaskschainofthought}, and GaoKaoBench \citep{zhang2024evaluatingperformancelargelanguage}.
Finally, we employ OpenCompass \citep{2023opencompass}, which is an LLM evaluation platform, supporting a wide range of models, to evaluate the results.
More details about training and evaluating can refer to Appendix~\ref{app:train and evaluate}.

\subsection{Main Results}
Table~\ref{tab:main} shows the main results of the two models' performances on five benchmarks after finetuned by CodeSeq.
We can summarize that:
(1) The sequence inductive reasoning synthetic data can effectively enhance the code generation capabilities of the two LLMs. After being finetuned with CodeSeq, the models achieve an average improvement of 3.67 points on Humaneval and 2.16 points on MBPP respectively.
(2) The sequence inductive reasoning synthetic data also demonstrates excellent transfer effects on comprehensive reasoning benchmarks (OOD). In particular, the LLaMA3-8B model improves by more than 8 points on MMLU. It is worth noting that although our CodeSeq data is in English, we still maintain the performance on the Chinese GaoKaoBench.

\begin{table}[t]
\renewcommand{\arraystretch}{1.2} 
 \setlength{\tabcolsep}{1.5mm}{
 \small
\begin{tabular}{|c|ccccc|}
\toprule
      & Heval & MBPP & MMLU & BBH & GK \\
\midrule
GPT4o & 92.70 &87.60 &88.70 &83.10&72.20 \\

\midrule
 LLaMA3-8B & 56.70 & 63.81 & 51.80 & 63.03 & 29.64 \\
 $+$ CodeSeq & 57.32 & 65.79 & 60.62 & 64.40 & 29.71 \\
  $\Delta$  & +0.62 & +1.98 & +8.82 & +1.37 & +0.07 \\
\midrule
 Qwen2.5-7B & 71.34 &71.59 & 68.23 & 66.05 & 63.29 \\
 $+$  CodeSeq & 78.05 & 73.93 & 70.74 & 69.70 & 63.77 \\
 $\Delta$  & +6.71 & +2.34 & +2.51 & +3.65 & +0.48  \\
\bottomrule
\end{tabular}}
\caption{Both models have improvements on five benchmarks, finetuned by CodeSeq. `Heval' and `GK' represent Humaneval and GaoKaoBench respectively.}
\label{tab:main}
\end{table}

\subsection{Ablation Study}
We conduct ablation studies with LLaMA3-8B.
From Table~\ref{tab:ablation}, we can conclude that:
(1) If Tulu3 is not used, the model will break down in terms of instruction-following ability, and the performances on various benchmarks will significantly decline.
(2) Training only with Tulu3 does not improve performance, so there will be no data leakage for the five benchmarks.
(3) The synthetic data will not improve compared to the original LLaMA3-8B if the case-based supervision signals are not injected.

\begin{table}[t]
\renewcommand{\arraystretch}{1.2} 
 \setlength{\tabcolsep}{1.5mm}{
 \small
\begin{tabular}{|c|ccccc|}
\toprule
      & Heval & MBPP & MMLU & BBH & GK \\
\midrule
 LLaMA3-8B & 56.70 & 63.81 & 51.80 & 63.03 & 29.64 \\
 $+$ CodeSeq & 57.32 & 65.79 & 60.62 & 64.40 & 29.71 \\
\midrule
 - Tulu3         & 49.68 & 56.33 & 54.14 & 60.26 & 23.18 \\
 - CodeSeq       & 54.65 & 61.72 & 51.91 & 62.88 & 28.45 \\
 - test cases     & 55.90 & 62.47 & 50.88 & 62.59 & 29.53 \\
\bottomrule
\end{tabular}}
\caption{We conduct ablation studies with LLaMA3-8B. `-Tulu3', `-CodeSeq' and `-test cases' mean only SFT with CodeSeq, only SFT with Tulu3 and deleting stage 3 in Figure~\ref{fig:pipline}, respectively.}
\label{tab:ablation}
\end{table}

\subsection{CodeSeq for Next Number Prediction}
\begin{figure}[t]
  \includegraphics[width=0.48\textwidth, height=0.15\textheight]{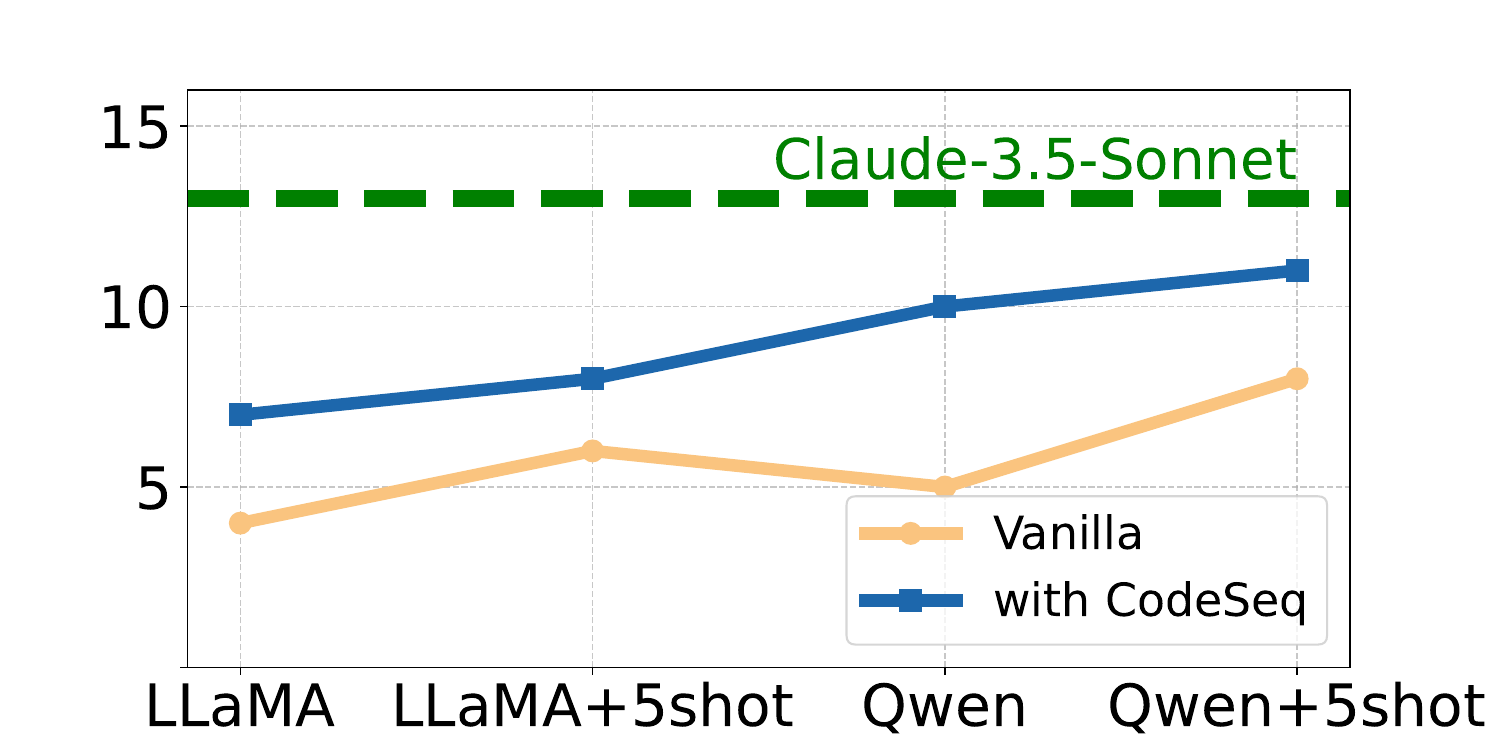}
  \caption{We respectively carry out next number prediction using LLaMA3-8B and Qwen2.5-7B before and after training, to test their inductive reasoning abilities.}
  \label{fig:next number2}
\end{figure}

We respectively carry out next number prediction using LLaMA3-8B and Qwen2.5-7B before and after training, to test their direct inductive reasoning abilities in Figure~\ref{fig:next number2}.
It can be concluded that in the 5-shot scenario, the accuracy of the models' prediction of the next number in a sequence will increase.
It is worth noting that the models trained with our CodeSeq reveal significant improvements in this task. 
Among them, after being trained with CodeSeq, Qwen2.5-7B's accuracy under the 5-shot setting is already close to the performance of Claude-3.5-Sonnet. 

\section{Conclusion}
In this paper, we novelly employ number sequences as the source of inductive reasoning data. 
To our knowledge, we are the first to utilize sequences as such kind of data to study their impact on LLMs.
We package the sequences into algorithmic problems, hence we can inject case-based supervision signals via code unit tests to improve data quality.
Our synthetic data CodeSeq is proven effective for various reasoning tasks, demonstrating the potential of inductive reasoning.

\section*{Limitations}
This paper takes sequences as a type of inductive reasoning data and explores the impact of this type of data on LLMs. 
We construct our own pipeline for generating synthetic sequence data and successfully combine it with code to insert process supervision signals. 
The finally formed CodeSeq training dataset is proven to have good effects on various reasoning tasks.
However, this article still has two limitations:
(1) Inductive reasoning tasks themselves are still in the initial stage of development. 
The significance of this type of task, the datasets, and the evaluation methods, etc., have not been systematically organized. Although we conduct preliminary explorations, this is a relatively novel direction and can be regarded as one of the future research works.
(2) Using only sequences as the data source for inductive reasoning is relatively limited. It is expected that more synthetic training data for inductive reasoning can be obtained in the future.

\section*{Ethics Statements}
The pipeline is primarily generated by deepseek-v3 and o1-preview. We obtain all the API Keys through a paid subscription.
The data source is the OEIS website, which is a public website.
The entire process and outcomes are free from intellectual property and ethical legal disputes.

\section*{Acknowledgments}
We will finish this part in the camera-ready version.
\bibliography{acl_latex}

\appendix

\section{Appendix}

\subsection{Next Number Prediction}
\label{app:next number}
Sequences are an excellent type of data for inductive reasoning because deriving the general term formula of a sequence requires inferring an abstract, universal representation based on the specific terms of the sequence.

After the process in the Sequence Synthetic Data Pipeline Section, we obtain sequence synthetic data. 
We randomly select 200 sequences and conduct the next number prediction experiments with the three most powerful LLMs in terms of reasoning ability: o1-preview, claude-3.5-sonnet, and deepseek-r1.
We ensure that these 200 test data are not used in the construction of CodeSeq.

\begin{figure}[ht]
  \includegraphics[width=\columnwidth]{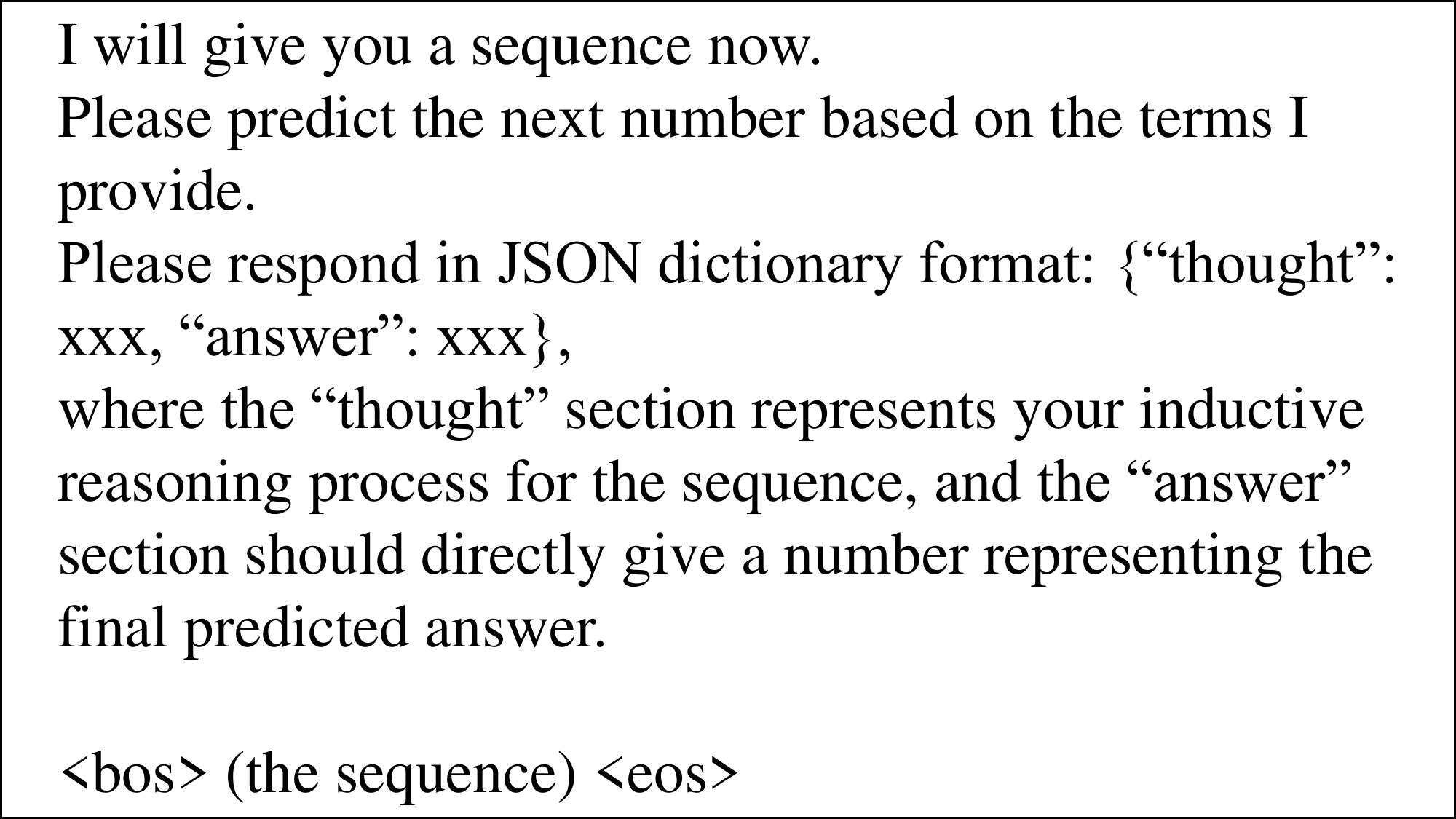}
  \caption{The prompt for the next number prediction task.}
  \label{fig:app_next_number}
\end{figure}

We use the following prompt in Figure~\ref{fig:app_next_number} to have it predict the next number in the given sequence.

\subsection{Related Work}
\subsubsection{Inductive Reasoning}
Reasoning can be mainly divided into two modes: deductive reasoning \citep{johnson1999deductive} and inductive reasoning \citep{hayes2010inductive}.
Deductive reasoning, such as well-defined tasks like mathematical and code reasoning \citep{DBLP:conf/iclr/WangRZLLSZSZ024,DBLP:journals/corr/abs-2410-08196}, utilizes general principles and axioms to achieve specific goals, pursuing logical certainty.
While inductive reasoning is quite the opposite.

Inductive reasoning, involving drawing general conclusions from specific patterns, is the most universal and essential method in knowledge discovery \citep{DBLP:journals/cogsr/HanRPK24}:
(1) Deriving general conclusions from specific cases, allowing it to cover and generalize to a wider range of applications, which aligns with the human learning process.
(2) Adaptive adjustments augment its reasoning ability in uncertain and complex scenarios, where inductive outcomes may not always be unique.

Despite its significance, existing works of LLMs reasoning are limited to deductive reasoning \citep{DBLP:conf/eacl/AhnVLLZY24,DBLP:journals/corr/abs-2403-00896,DBLP:conf/nips/LiuXW023,DBLP:journals/tosem/JiangDWFSLJJ24}. 
This is because obtaining high-quality process supervision data is quite challenging for inductive reasoning \citep{yang2024languagemodelsinductivereasoners}.
So this paper aims to overcome such limitation.

\subsubsection{Code Reasoning}
Code serves as a crucial link between humans and machines. It is ultimately converted into specific programs that can replace human labor in fulfilling diverse tasks. These programs are marked by several notable traits, including precision, logical structure, modular design, and excitability \citep{wan2023deeplearningcodeintelligence,sun2025surveyneuralcodeintelligence}.

In the era of AI, code generation mainly consists of three stages: (1) the code embedding \citep{DBLP:conf/cvpr/GirdharELSAJM23}, (2) code pre-trained models \citep{DBLP:journals/ijautcomp/WangCQGWWTG23}, and (3) code generation in LLMs. These three stages have corresponding relationships with the development of natural language processing.

The most prominent feature of code generation is learning with execution feedback \citep{DBLP:conf/nips/YangPNY23}. 
Code has an inherent property of being compliant and executable. This enables compilers or interpreters to automatically produce accurate feedback.
This process can be called the code unit tests \citep{DBLP:conf/nips/Le0GSH22}.

In the era of LLMs, there are three main methods for enhanced code generation:
(1) Decoding-enhanced, that is, using methods such as self-planning \citep{jiang2024selfplanningcodegenerationlarge} and self-filling \citep{martinez2023hierarchical}, and guiding the generation of code in combination with the Program of Thought (PoT) \citep{DBLP:conf/aaai/Bi0JDZC24} technology.
(2) Feedback-drive, which is similar to tree search \citep{DBLP:journals/pacmpl/MatuteNBCC24,DBLP:conf/nips/DaineseMAM24} and uses unit tests to provide supervision signals.
(3) Natural-language (NL) guidance \citep{DBLP:conf/issta/WangGZSSZZS0X24}, that is, using natural language to guide the generation of code.

In this paper, we explore injecting case-based code supervision signals to improve inductive reasoning data quality.

\subsection{The Sequence Inductive Reasoning Synthetic Data Pipeline}
\label{app: pipeline}
In this section, we will provide more detailed information and more examples to clearly explain the sequence synthetic data pipeline.
For the working agent, considering that we need to make frequent calls, and for cost-saving purposes, we chose deepseek-v3\footnote{https://www.deepseek.com/} \citep{deepseekai2024deepseekv3technicalreport}, while for the guiding agent, we select the currently most powerful reasoning model, o1-preview\footnote{https://openai.com/o1/}, so that the self-correction process will be more accurate.
We will demonstrate how these strong instructions-following agents work under the guidance of prompts with detailed instructions.

\subsubsection{Sequence Data Filtering}
\begin{figure*}[t]
  \includegraphics[width=\linewidth]{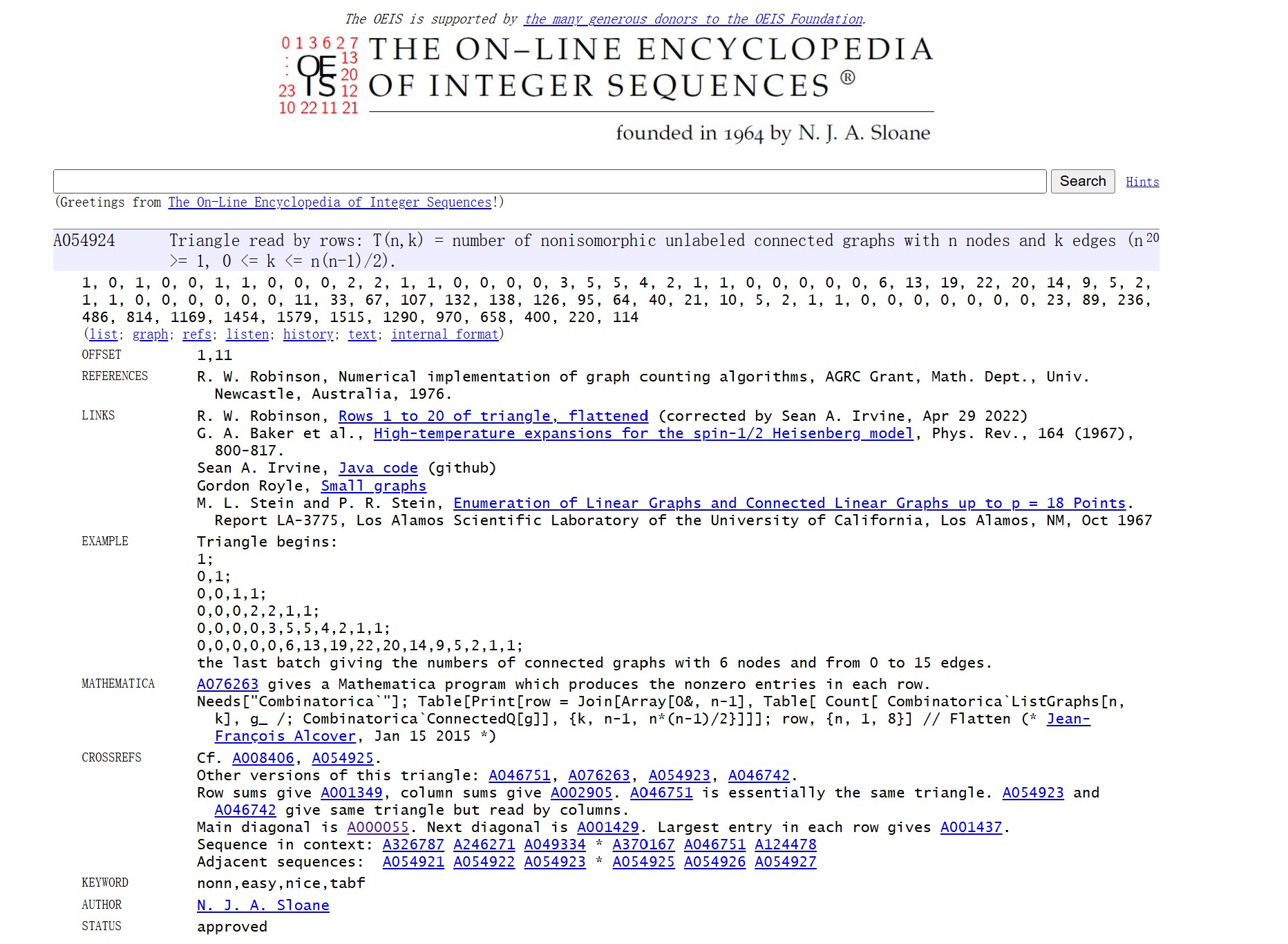}
  \caption{An example of one OEIS webpage. This webpage includes the sequence, sequence offsets, sequence references, sequence links to other supplementary information, examples in the explanation process, mathematical explanations, the relationship between sequences, and so on.}
  \label{fig:app_oeis}
\end{figure*}

We scrape a large number of sequences and their related information from the OEIS website. 
Each page on the website corresponds to a sequence and all its information, including the source, formula, general term description, and so on.
We give an example of one OEIS webpage in Figure~\ref{fig:app_oeis}.

\begin{figure*}[ht]
  \includegraphics[width=\linewidth]{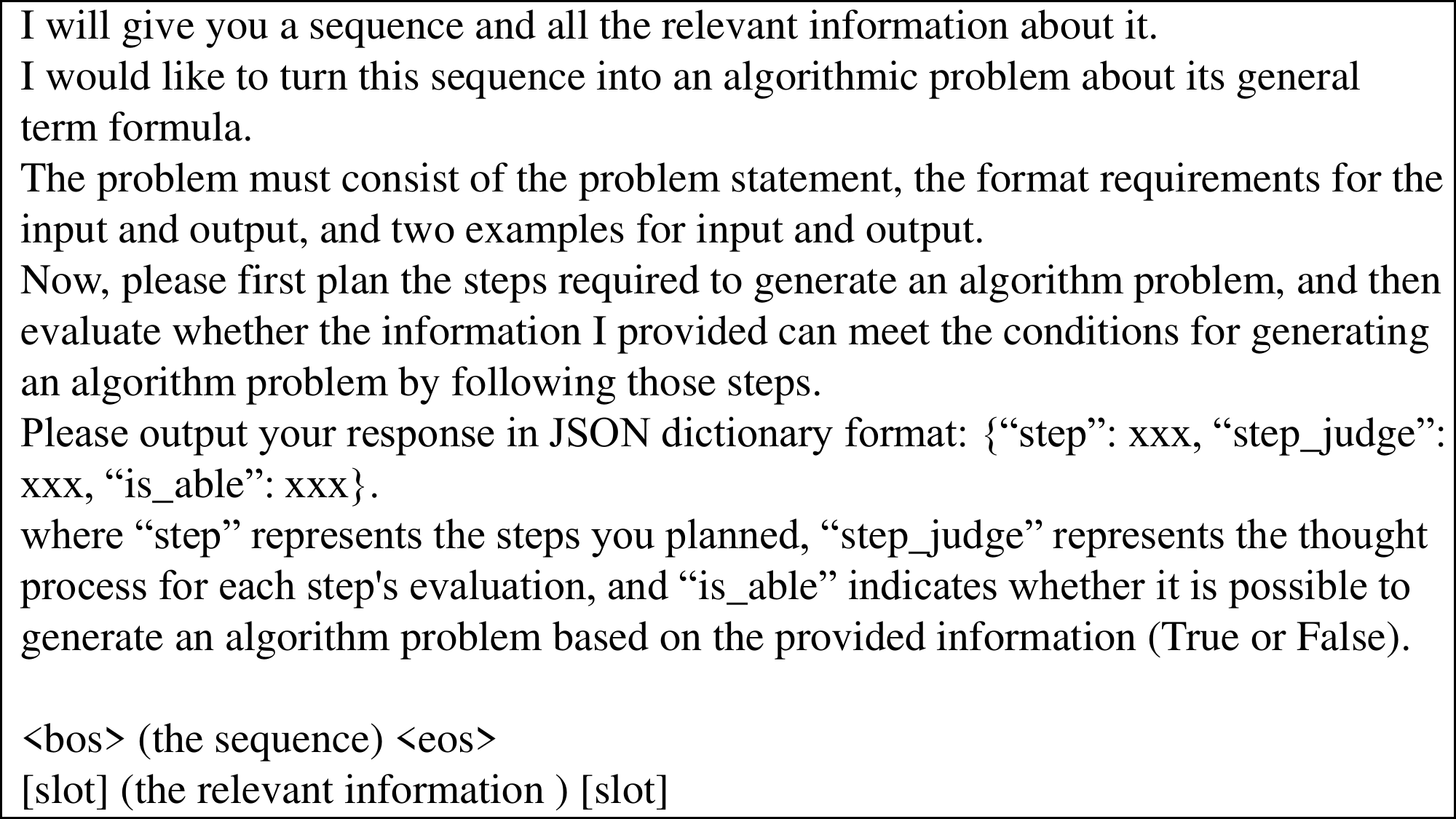}
  \caption{The prompt for the working agent to conduct self-planning on the problem generation and self-reflecting on whether each step contains enough information.}
  \label{fig:prompt_enough_information}
\end{figure*}

We need to filter the information for each candidate sequence to ensure the accuracy of the algorithmic problem generation process.
We first manually wrote rules to filter out sequences with insufficient information, including:
(1) those with too few terms, which will result in any powerful agent being unable to thoroughly understand the mathematical logic of the sequence.
(2) those that evolve from other sequences, which will result in us being unable to crawl enough information about the current sequence from the existing website.
(3) those without "mathematical" or "programming" fields, this is for the working agent to initially filter information, making it easier to generate algorithm problems.
Then we prompt the working agent to self-planning the steps for generating an algorithmic problem and self-reflecting on whether each step contains enough information.
This prompt are shown in Figure~\ref{fig:prompt_enough_information}.
The above operations result in a batch of sequences with high information density.

\subsubsection{Sequence Algorithmic Problem Generation and Validation}
\begin{figure*}[ht]
  \includegraphics[width=\linewidth]{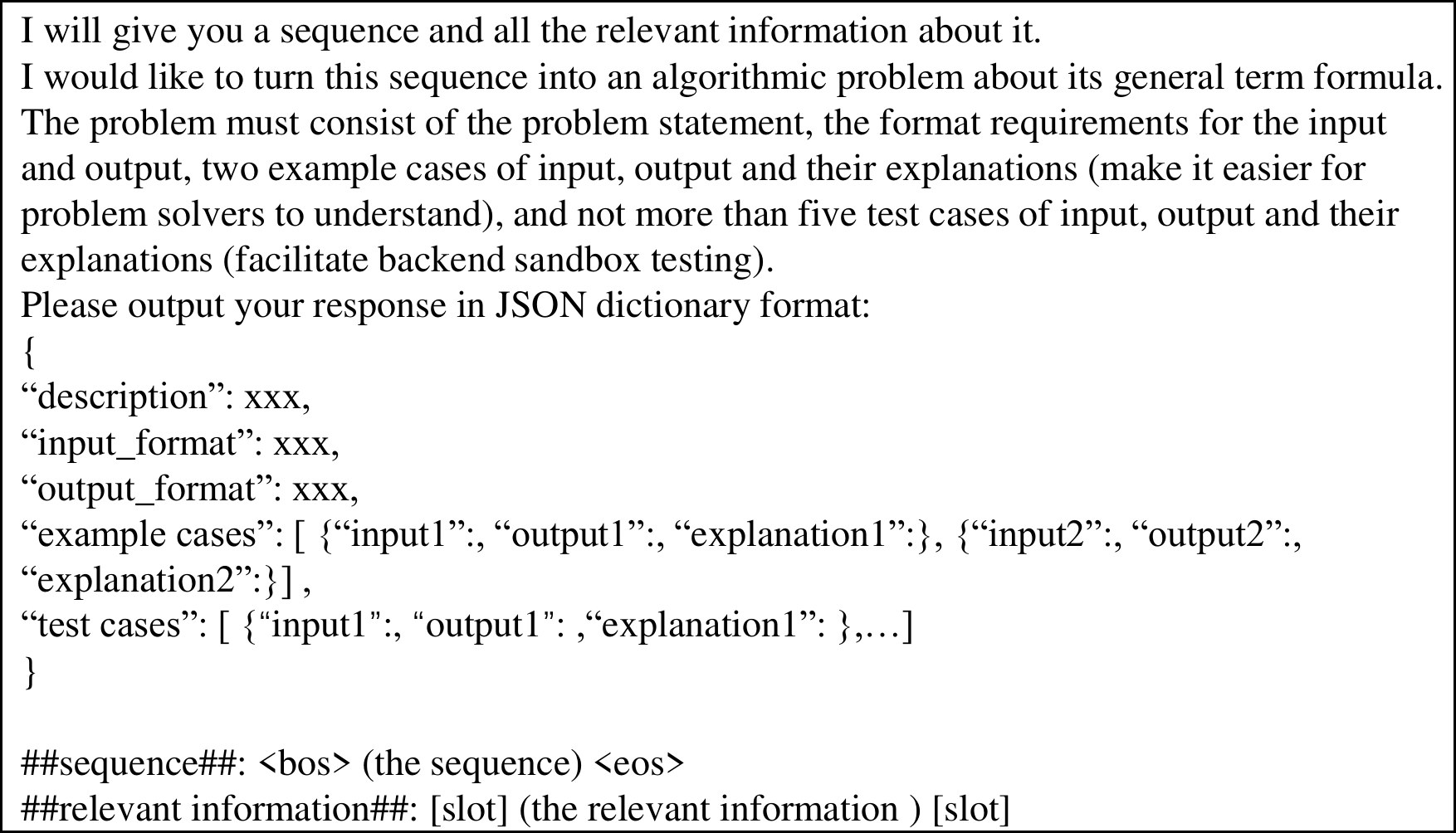}
  \caption{The prompt for algorithmic problem generation.}
  \label{fig:problem_generation}
\end{figure*}

\begin{figure*}[ht]
  \includegraphics[width=\linewidth]{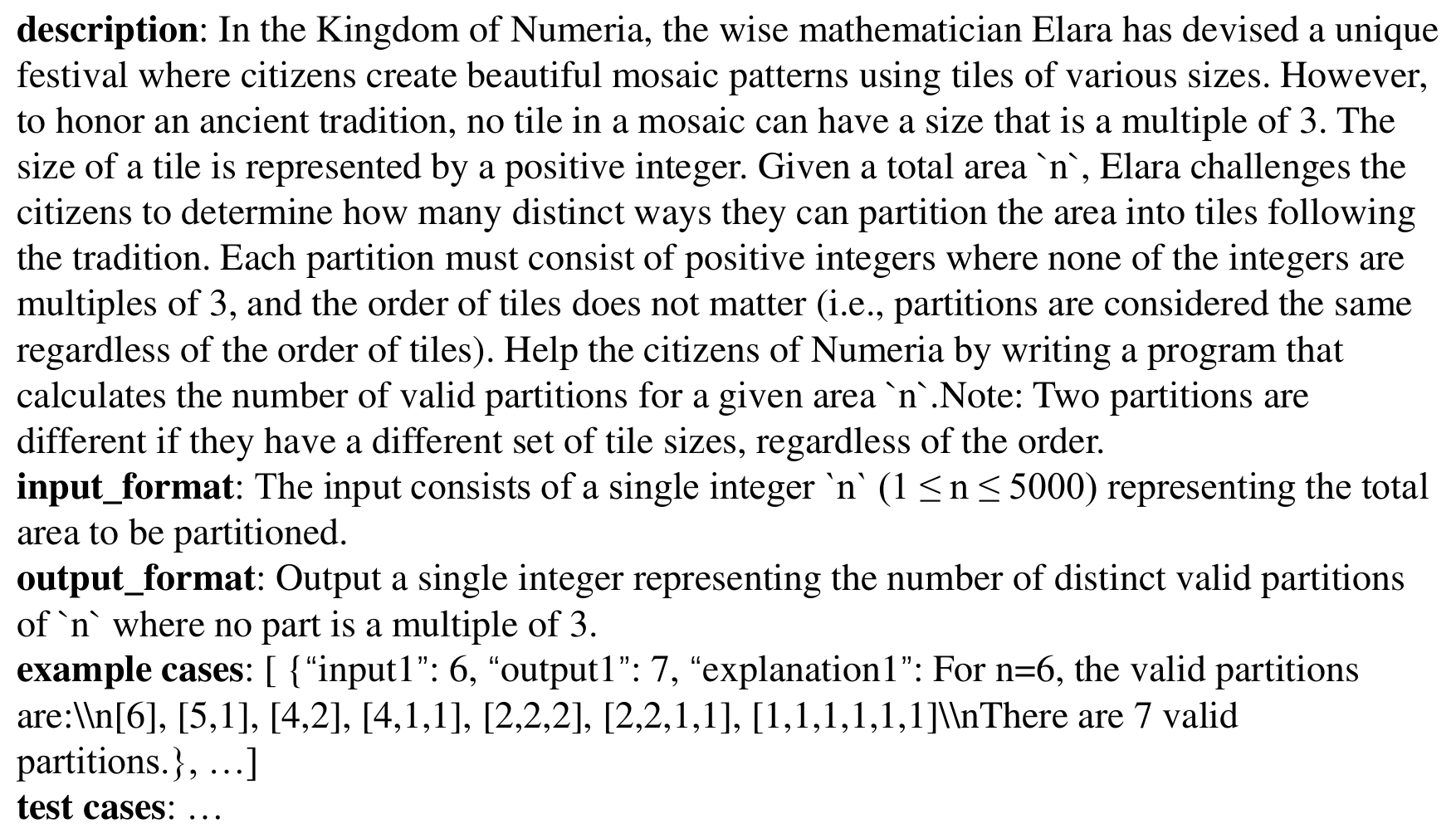}
  \caption{A generated example for one sequence algorithmic problem.}
  \label{fig:generation_exapmle}
\end{figure*}

We next have the working agent generate an algorithmic problem about the general terms for each sequence, along with two example cases. 
The prompt for problem generation is in Figure~\ref{fig:problem_generation}.
Example cases provide the standard input and output cases for this algorithmic problem to help the problem solvers understand it better.
We also give a generated example in Figure~\ref{fig:generation_exapmle}.

\begin{figure*}[ht]
 \centering
  \includegraphics[scale=0.3]{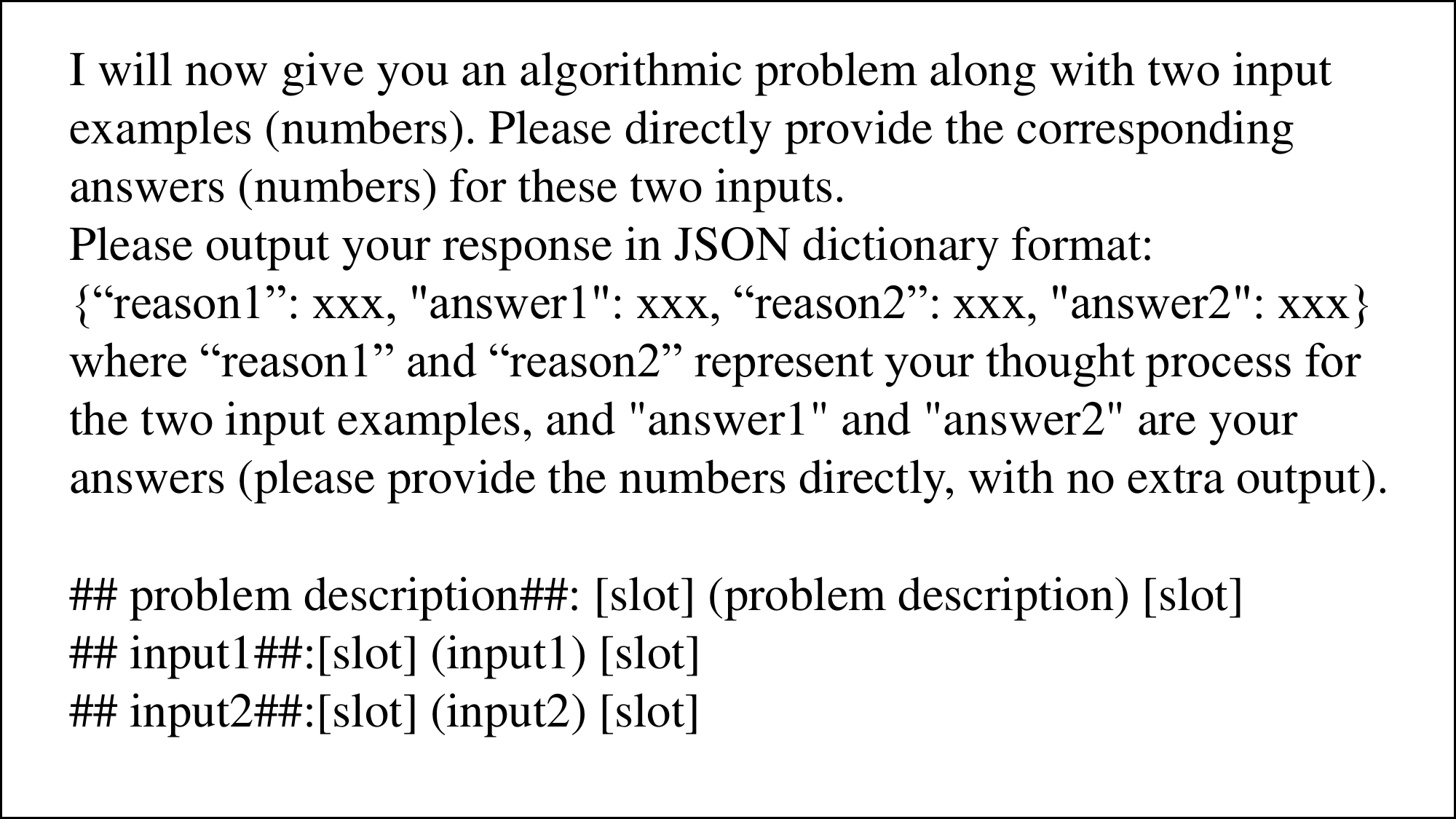}
  \caption{The prompt for the guiding agent directly outputs the results so that we can determine whether the current problem is correct.}
  \label{fig:direct_output}
\end{figure*}

To further verify the correctness of the algorithmic problems, we utilize another powerful LLM as a guiding agent. 
We input the problem description and two example cases' inputs into it and let it directly output the results (prompt in Figure~\ref{fig:direct_output}). 
By comparing these outputs with the ground truth outputs generated by the working agent, we can determine whether the current problem is correct.
Seed sequence data is gained via this example case validation.
Take the algorithmic problem in Figure~\ref{fig:generation_exapmle} as an example, if the guiding agent outputs 7 for the first example case, it matches the ground truth. 
If both the answers match the ground truth in example cases, we can say that the current generated problem is correct.

\subsubsection{Case-based Supervision Signal Injection}
\begin{figure*}[ht]
  \centering
  \includegraphics[scale=0.3]{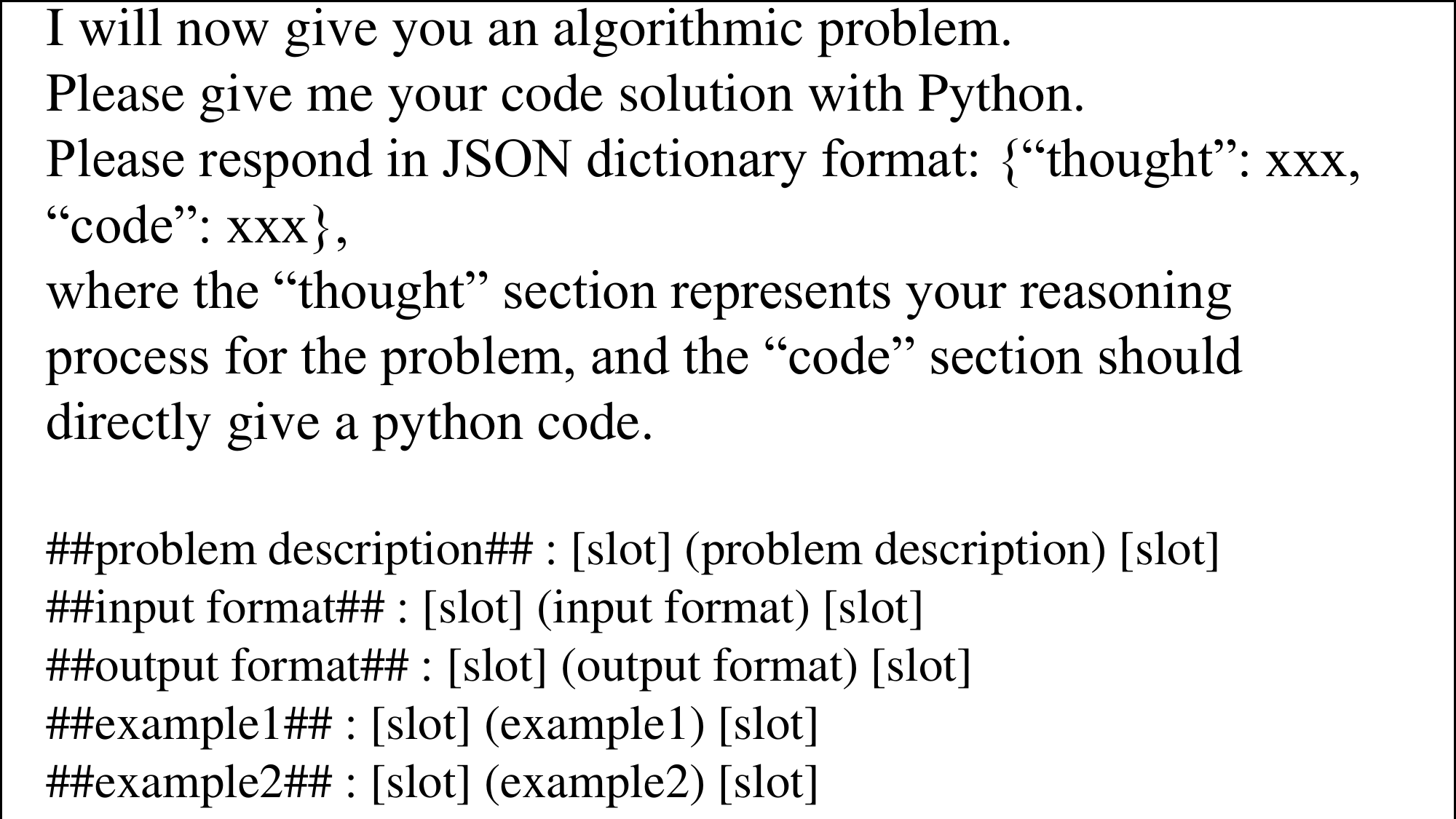}
  \caption{The prompt for the first time code solution generation.}
  \label{fig:code_generated1}
\end{figure*}

After obtaining the seed data, we let the working agent directly generate the code solution for the algorithmic problem with the prompt in Figure~\ref{fig:code_generated1}.

Since the problem description involves the general term of a sequence, the code solution represents the computational process for the general term of the sequence.
Unlike the example cases, we also set 5 to 7 test cases for each sequence to ensure the correctness of the code solution, as illustrated in Figure~\ref{fig:problem_generation}.

Imitating previous unit tests \citep{hui2024qwen25codertechnicalreport}, we use test cases to test the correctness of each code solution in an isolated sandbox environment.
A sandbox environment for executing code \citep{DBLP:conf/usenix/LiMNPBD14,DBLP:journals/jamia/CohnKMJPB24} is a controlled and isolated setting where code can be run without affecting the host system or other applications. In this environment, the code is executed within a restricted space, preventing it from accessing sensitive resources, files, or system-level operations outside the sandbox. Sandboxes are commonly used for testing, experimentation, and security purposes, as they allow developers to execute potentially untrusted or experimental code safely. The goal is to mitigate risks, such as malware or unintentional system damage, by containing the code's actions and ensuring it can not interfere with critical parts of the system.
Our code sets up a sandbox environment to safely execute user-provided Python code. It isolates the code by removing access to potentially dangerous built-in functions like open, exec, and eval, and replaces the print function with a safe version. We also redirect input and output to custom streams to capture them. The code is executed in a controlled environment with only a limited set of built-in functions available. If errors occur, they are caught and formatted with details, including the line number. Finally, we restore the system’s original state after execution. This approach ensures safe, isolated execution of potentially risky code.

\begin{figure*}[ht]
  \includegraphics[width=\linewidth]{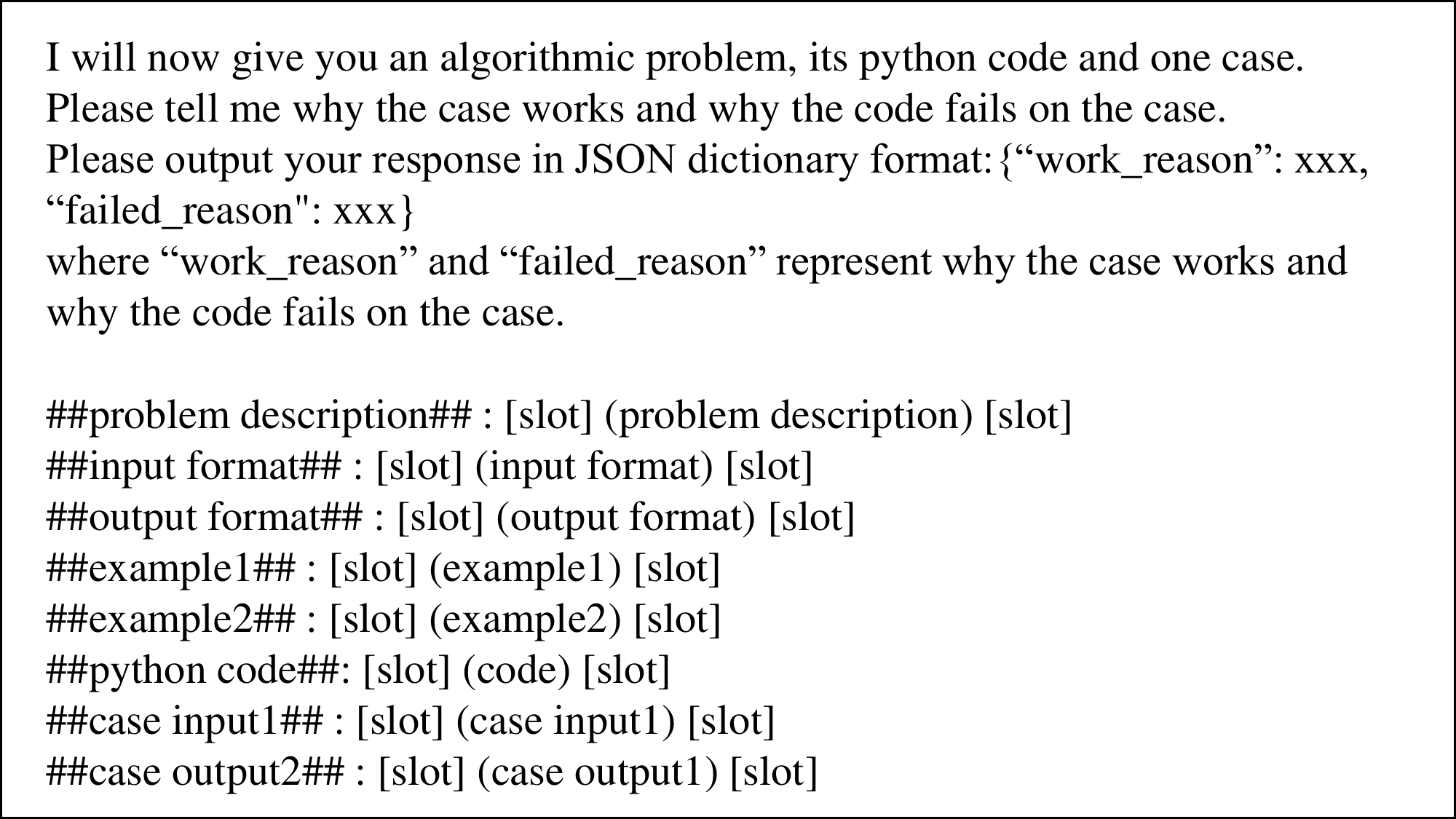}
  \caption{This prompt is inputted into the guiding agent to generate the reason why such case fails on the current code solution.}
  \label{fig:give_reason}
\end{figure*}

\begin{figure*}[ht]
  \includegraphics[width=\linewidth]{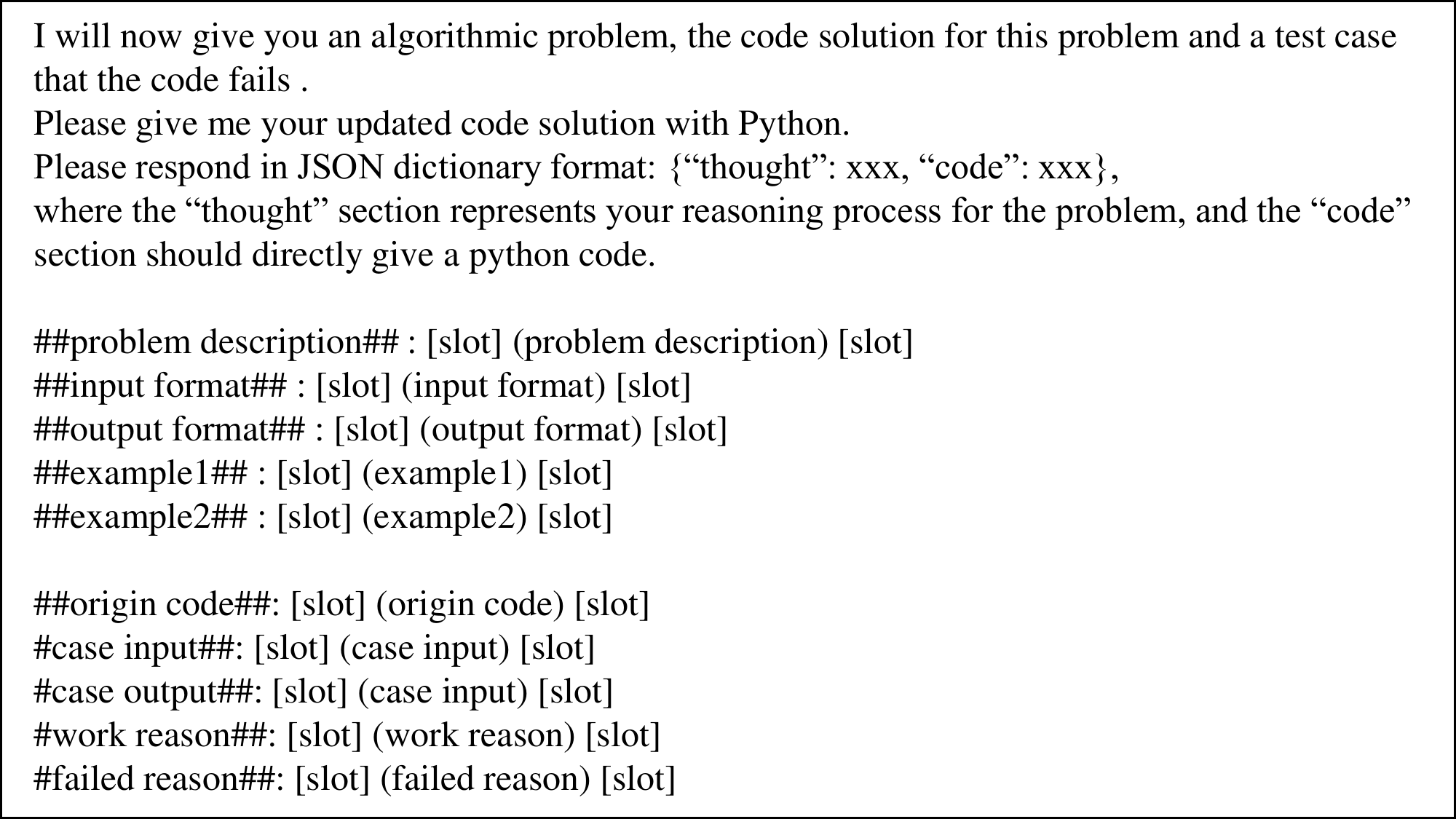}
  \caption{The prompt for the working agent to regenerate and correct the code.}
  \label{fig:code_generated2}
\end{figure*}

If a code solution fails on a test case, we ask the guiding agent to provide the reason for the failure (Figure~\ref{fig:give_reason}).
We then give that reason along with the test case back to the working agent to correct the code solution.
The prompt for the working agent to regenerate and correct the code is in Figure~\ref{fig:code_generated2}.
Ultimately, through continuous self-correcting, we achieve a code solution that passes all the test cases.

\subsection{The CodeSeq Dataset}
\label{app:statistical}
Based on the above process, we record the code of the current version each time a modification is made and generate synthetic data for each sequence, then form a training dataset CodeSeq.

\begin{figure*}[ht]
  \includegraphics[width=\linewidth]{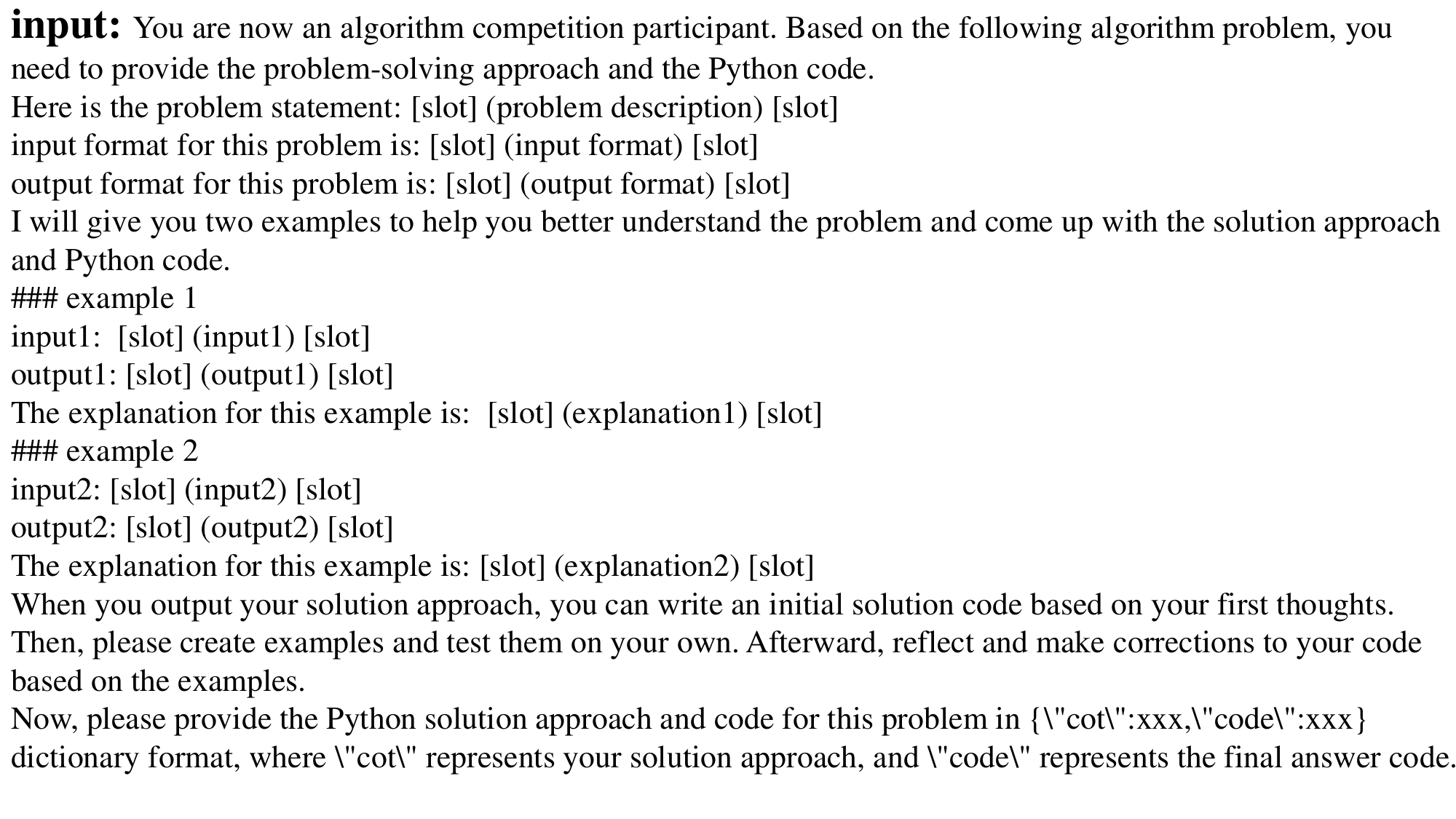}
  \caption{The standard input of one sft training data in CodeSeq.}
  \label{fig:sysdata_input}
\end{figure*}

\begin{figure*}[ht]
  \includegraphics[width=\linewidth]{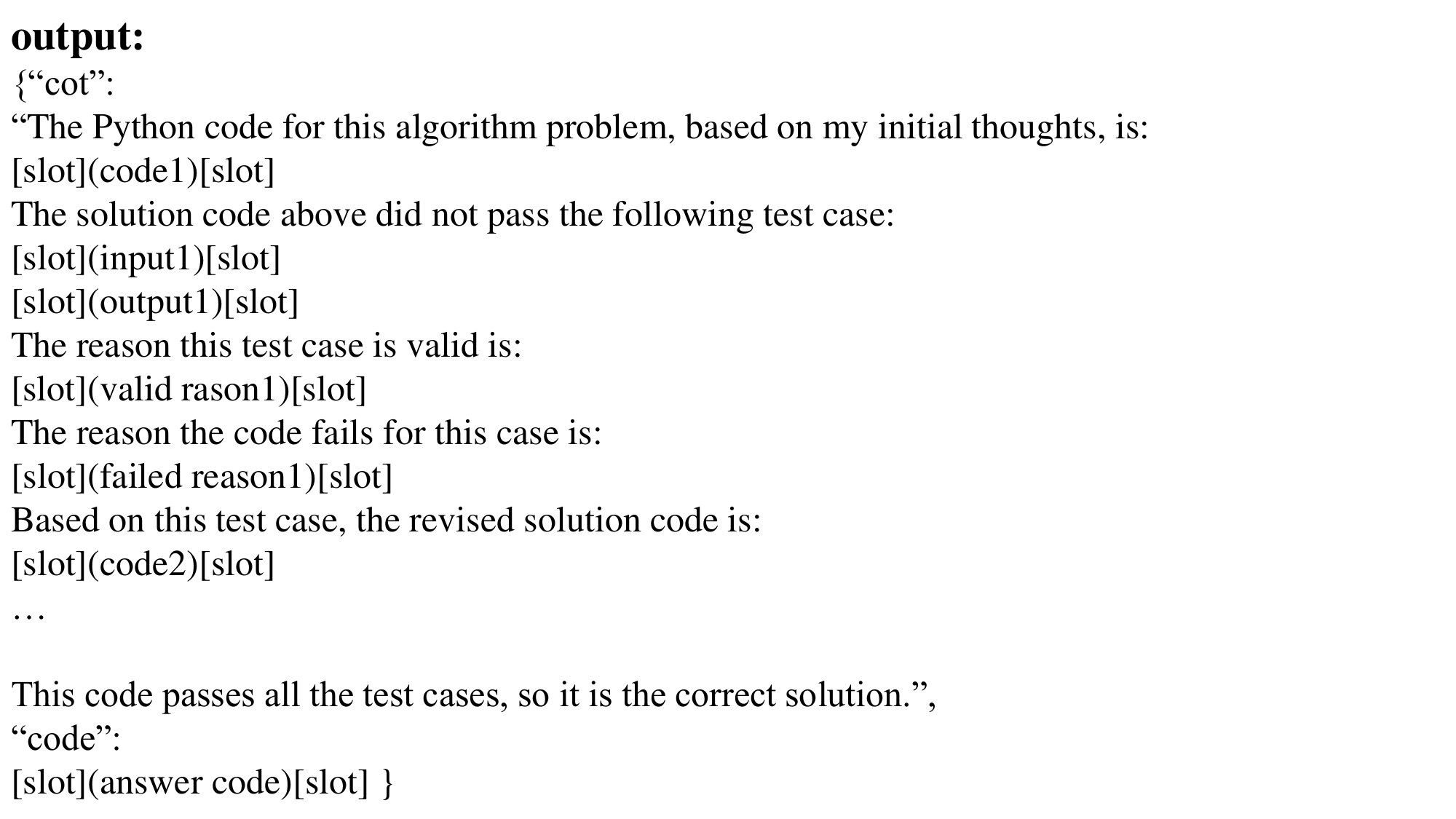}
  \caption{The standard output of one sft training data in CodeSeq.}
  \label{fig:sysdata_output}
\end{figure*}

Our training data is primarily used for model training in the post-trained stage (especially SFT), so our dataset is organized in the SFT format.
A standard SFT input format in CodeSeq is shown in Figure~\ref{fig:sysdata_input}, and a standard SFT output format in CodeSeq is shown in Figure~\ref{fig:sysdata_output}.
As with other powerful reasoning models, we use the Chain-of-Thought (CoT) technique \citep{DBLP:journals/tse/YangZCZZC24} to guide the model’s deep reasoning process. 
In the output format, we store the CoT field and the final answer field separately.

\subsection{Details for Training and Evaluation}
\label{app:train and evaluate}
\subsubsection{LLM Backbones}
We conduct SFT on two widely used LLMs: LLaMA3-8B-Instruct and Qwen2.5-7B-Instruct.

\paragraph{LLaMA3-8B-Instruct} \citep{grattafiori2024llama3herdmodels}
LLaMA3-8B is an advanced LLM developed by Meta, featuring 8 billion parameters. It is part of the Llama 3 family.
This model is built on an optimized Transformer architecture and trained on a diverse dataset of over 15 trillion tokens. The training dataset includes a significant amount of code and covers over 30 languages, with more than 5\% of the data being non-English.
LLaMA3-8B is particularly designed to excel in instruction-based tasks, making it highly effective for scenarios requiring precise and context-aware responses.

\paragraph{Qwen2.5-7B-Instruct} \citep{qwen2025qwen25technicalreport}
Qwen2.5-7B is a powerful LLM developed by Alibaba's ModelScope team, featuring 7.6 billion parameters. It is designed to excel in various natural language processing tasks, with notable strengths in long-context understanding, multilingual support, and specialized capabilities for coding and mathematical tasks.
This model supports up to 128K tokens for context understanding and can generate up to 8K tokens of text, making it highly effective for long-text generation and structured data processing. What's more, Qwen2.5-7B is trained on a massive 18T dataset.

\subsubsection{Mix Training Details}
To maintain the models’ instruction-following ability, we mix CodeSeq with the latest post-training dataset Tulu3 \citep{lambert2025tulu3pushingfrontiers} for SFT.

\paragraph{Tulu3} is a comprehensive dataset and training framework developed by the Allen Institute to advance the post-training of LLMs. The Tulu3 dataset is designed to enhance language models' performance through SFT and reinforcement learning. It includes a mixture of data from various sources, covering a wide range of natural language processing tasks such as instruction following, mathematical reasoning, and code generation.

Due to the timeliness of Tulu3, we ensure that it is not used for any backbone model training. During the training process, we removed samples longer than 5120 tokens and excluded all samples related to mathematics and code (since we focus on code and comprehensive reasoning tasks). Finally, we retain over 800k training samples of Tulu3.

To improve the models' reasoning ability while maintaining its other capabilities, particularly instruction-following ability, we calculate the average number of tokens in the Tulu3 and CodeSeq datasets. We assign a weight ratio of 5:1 to these two datasets for mixed training. During training, we wrap all inputs and outputs with chat templates to prevent the loss of instruction-following capabilities.

\subsubsection{Training Parameters}
We conduct SFT on two widely used LLMs: LLaMA3-8B and Qwen2.5-7B based on InternTrainer\footnote{https://github.com/interntrainer} framework with 8 NVIDIA-L20Y.
The training parameters are shown in Table~\ref{tab: training parameters}.

\begin{table}[ht]
\centering
\fontsize{9pt}{9pt}\selectfont
\renewcommand{\arraystretch}{1.5} 
\begin{tabular}{cc}
\toprule
total-steps                 & 1000                 \\
epochs                      & 1                    \\
bzs                         & 16                   \\
gradient-accumulation       & 16                   \\
micro-bsz                   & 1                    \\
seq-len                     & 5120                  \\
max-length-per-sample       & 5120                  \\
min-length                  & 50                  \\
num-worker                  & 4                    \\
loss-label-smooth           & 0                    \\
lr                          & 1e-5                    \\
warmup-ratio                & 0.1                    \\
weight-decay                & 0.01                   \\
\midrule
adam-beta1                  & 0.9                   \\
adam-beta2                  & 0.95                   \\
adam-eps                    & 1e-8                   \\
\midrule
fp16-initial-scale          & 2**14                    \\
fp16-min-scale              & 1                    \\
fp16-growth-interval        & 1000                    \\
fp16-growth-factor          & 2                   \\
fp16-backoff-factor         & 0.5                   \\
fp16-max-scale              & 2**24                   \\
\midrule
zero1-size                  & 8                      \\
tensor-size                 & 1                      \\
pipeline-size               & 1                      \\
weight-size                 & 1                      \\
\bottomrule
\end{tabular}
\caption{The training parameters.}
\label{tab: training parameters}
\end{table}

\subsubsection{Benchmarks}
We test the tuned models on two code benchmarks: Humaneval \citep{chen2021codex} and MBPP \citep{austin2021programsynthesislargelanguage}, along with three comprehensive reasoning benchmarks: MMLU \citep{hendrycks2021measuringmassivemultitasklanguage}, BBH \citep{suzgun2022challengingbigbenchtaskschainofthought}, and GaoKaoBench \citep{zhang2024evaluatingperformancelargelanguage}.

\paragraph{Humaneval} consists of 164 hand-crafted programming challenges that are comparable to simple software interview questions, each with a function signature, natural language description, and unit tests to validate the correctness of generated code.

\paragraph{MBPP} The MBPP (Mostly Basic Python Problems) benchmark consists of around 1,000 crowd-sourced Python programming problems, each with a task description, code solution, and three automated test cases.

\paragraph{MMLU} The MMLU (Massive Multitask Language Understanding) benchmark is a comprehensive evaluation tool designed to assess the knowledge and reasoning capabilities of LLMs across a wide range of academic and real-world subjects.

\paragraph{BBH} The Big Bench Hard (BBH) benchmark is a collection of challenging tasks designed to evaluate the reasoning and logical abilities of LLMs.

\paragraph{GaoKaoBench} The GAOKAO-Bench is an evaluation framework that uses Chinese college entrance examination (Gaokao) questions as its dataset to assess the language understanding and logical reasoning capabilities of LLMs. It includes a comprehensive collection of questions from 2010 to 2023.
For convenience in evaluation, we select only objective questions for testing.

\subsubsection{Compared Models}
We chose GPT4o as the target baseline because it undergoes the most systematic evaluations across various benchmarks. Since the parameters in our models are much fewer than that of GPT4o, it is difficult for our model to outperform it in all aspects. Nevertheless, we can still present the corresponding results. This will facilitate everyone's understanding of the gap and motivate us to strive for catching up.

\paragraph{GPT4o} GPT4o\footnote{https://openai.com/index/hello-gpt-4o/} is an advanced AI model that provides more accurate and efficient language processing capabilities by OpenAI. It builds upon the strengths of previous models while incorporating new optimizations to enhance performance. With its ability to understand and generate human-like text, GPT4o aims to assist users in various tasks such as writing, problem-solving, and information retrieval.

\subsubsection{OpenCompass}
We employ OpenCompass\footnote{https://github.com/open-compass/opencompass} \citep{2023opencompass}, which is an LLM evaluation platform, supporting a wide range of models, to evaluate the results.
It features a wide range of capabilities, including language understanding, reasoning, coding, and long-text generation, and provides a fair and reproducible benchmark for model evaluation.

We apply the Hugging Face framework to infer the models.
For code generation, the settings are: \textit{\{max-out-len: 1024, max-seq-len: 2048, batch-size:4, min-new-tokens: 50, num-return-sequences: 1, top-p: 0.9, num-beams: 10\}}.
For other generation, the settings are: \textit{\{max-out-len: 1024, batch-size:8, min-new-tokens: 10\}}

For Humaneval, MBPP, and comprehensive reasoning benchmarks, we use pass@1, pass score, and selection accuracy as metrics separately. We ensure that all baselines are tested with the same settings for a fair comparison.

\end{document}